\documentclass[a4paper,11pt]{article}

\usepackage{fourier}
\usepackage{color}
 \usepackage{graphicx}
\usepackage{url}
\usepackage[affil-it]{authblk}
\usepackage{amsmath}
\usepackage{wrapfig}
\usepackage{natbib}

\usepackage[T1]{fontenc}
\usepackage{times}

\usepackage{caption}
\usepackage{subcaption}

\begin{document}

\title{A 2.5D Vehicle Odometry Estimation for Vision Applications}

\author[1]{Paul Moran \thanks{paul.moran@valeo.com}}
\author[2]{Leroy-Francisco Periera}
\author[2]{Anbuchezhiyan Selvaraju}
\author[2]{Tejash Prakash}
\author[1]{Pantelis Ermilios}
\author[1]{John McDonald}
\author[1]{Jonathan Horgan}
\author[1,3]{Ciar\'{a}n Eising}
\affil[1]{Valeo Vision Systems, Dunmore Road, Tuam, Co. Galway, Ireland}
\affil[2]{Valeo India, Rajiv Gandhi Salai, Navalur, Chennai - 600 130, Tamil Nadu, India}
\affil[3]{Department of Electronic and Computer Engineering, University of Limerick, Ireland}
\date{}
\maketitle
\thispagestyle{empty}

\vspace{-1cm}

\begin{abstract}
This paper proposes a method to estimate the pose of a sensor mounted on a vehicle as the vehicle moves through the world, an important topic for autonomous driving systems. Based on a set of commonly deployed vehicular odometric sensors, with outputs available on automotive communication buses (e.g. CAN or FlexRay), we describe a set of steps to combine a planar odometry based on wheel sensors with a suspension model based on linear suspension sensors. The aim is to determine a more accurate estimate of the camera pose. We outline its usage for applications in both visualisation and computer vision.  
\end{abstract}

%%%%%%%%%%%%%%%%%%%%%%
\section{Introduction}

%%%%%%%%%%%%%%%%%%%%%%
A real-time estimate of a vehicle's pose in a world coordinate system is important for Advanced Driver-Assistance Systems (ADAS) and autonomous vehicles. %Examples include path planning for automated parking \citep{brunker2018}, automated lane change \citep{peter2019}, and vehicle autonomy in general \citep{geiger2012}. 
Accurately measuring the pose of sensors attached to the vehicle is also vital for perception. Using accurate 3D odometry, the task of finding the sensor pose with known rigid body extrinsics within the world coordinate system is trivial through a simple coordinate system change. However, limited to only 2D odometry the pose of the sensor may be inaccurate due to suspension changes which are unaccounted for in 2D (planar) odometry.
For instance, if there is a heavy loading in the vehicle, causing a suspension change from the nominal, then the system will have an inaccurate estimate of the camera extrinsics. This will cause an issue for any vehicular mapping system as without an accurate sensor ego pose the system cannot accurately localise perceived objects from that sensor relative to the vehicle.

Vehicle odometry %(2D and 3D)
can be estimated from various sensor types. Laser scanners can be used to estimate 2D odometry \citep{jaimez2016}. Despite their accurate odometry estimates, due to their expense they are not universally deployed in vehicles. Visual odometry remains a significant area of research, and though it can give very high accuracy, the issue of scale resolution is still an unsolved topic \citep{liu2018}. 
High-grade Global Navigation Satellite Systems (GNSS) and Inertial Navigation Systems (INS) can offer greater accuracy than wheel-based odometries \citep{aqel2016}, but again are expensive and as such fail with ubiquitous deployment on vehicles \citep{gonzalez2019}. Visual-Inertial Odometry \citep{scaramuzza2019} is a method that combines visual and inertial sources of odometry to overcome limitations of both sensor types. However, it suffers from the same problems of universal deployment as INS. Hence, wheel-based odometry remains popular, and continues to be an area of research in robotics and autonomous vehicles \citep{brunker2018}. For further information, the reader is referred to \cite{mohamed2019}, who give a very complete overview of odometry in autonomous systems.
%,toledo2018}.

Wheel encoders are commonly deployed on vehicles \citep{brossard2019} where the sensor information is broadcast on the vehicle's system bus (CAN, FlexRay, Ethernet). Typically, these encoders utilise Hall effect sensors \citep{popovic2003}
%, ramsden2006}
though research continues into potentially better alternatives \citep{shah2019}. %One sensor is deployed for each wheel. 
To detect changes in heading, two common sensor types are deployed in vehicles: steering angle sensors (e.g. rotary potentiometer \citep{todd1975}) and/or yaw rate sensor (e.g. gyroscope \citep{passaro2017}).

%\textcolor{blue}{Tracking the position of a sensor located on the body of a vehicle is essential for optical flow, and Simultaneous Logicalisation and Mapping (SLAM) approaches}. 
%\citet{mariotti2019} used wheel odometry to estimate the motion of cameras on a vehicle, and use this motion in combination with optical flow to segment motion in scenes in which the vehicle itself is under motion. 
Traditionally, wheel-based odometry was only used to provide a planar motion estimate of the vehicle; calculating an odometry estimate with only three degrees of freedom. Here we propose to augment the planar 2D wheel odometry by using sensors that measure the current level of the suspension of the vehicle using linear potentiometers \citep{todd1975}, giving a much better estimate of the true extrinsic position of the camera for a given moment in time. These advanced sensors are becoming commonplace on vehicles with adjustable suspension for altering ride height. % on which they are equipped. 
The technique described does not give a full 3D odometry estimate, but could metaphorically be referred to as a 2.5D estimate of odometry of the vehicle. In this paper, we use the yaw rate sensor, as this enables us to avoid using a specific model of vehicle steering (e.g. Ackermann), which may have inaccuracies. %Finally, in some modern vehicles, changes in suspension are reported using linear potentiometers \citep{todd1975}, which are calibrated to report the height of the wheel arch from the ground. 

This paper is organised as follows. In the following section, we discuss the motion of the vehicle on the ground plane, and the motion of the sensors due to changing suspension, and how both can be combined. In Section \ref{sec:results}, we provide some results, examining the accuracy of the planar odometry, the behaviour of the sensors, and some results in application for human visualisation and computer vision.

\section{Proposed Method}

%\subsection{Note on coordinate systems and notation}
We define the coordinate system of the vehicle to have the origin at the rear axle, $X^v$-axis pointing forward in the direction of the vehicle, $Z^v$-axis pointing upward, roughly orthogonal to the ground plane, and $Y^v$-axis in the direction of left hand turning ($X^v$-axis is shown in Figure \ref{fig:ackerman1}). We track the position of the vehicle in a world coordinate system with axes $X^w$, $Y^w$ and $Z^w$. % (Figure \ref{fig:ackerman2}). 
%We assume no external source of position estimation of the vehicle (e.g. GPS), so typically the world coordinate system will be coincident with the first iteration of the vehicle coordinate system. 
For vectors, we use the super-scripts $v$, $w$ and $c$ to indicate the coordinate system in which the vector is defined: the vehicle, world and sensor (typically camera) coordinate system. We use $\theta(t)$ to denote a continuous function in $t$ and $\theta'(t)$ to denote it's derivative with respect to $t$. When using sampled data, we use numbered subscripts instead of parentheses, e.g. $\theta'_1$, $\theta_1$, $t_1$, etc. In this paper, we don't discuss the uncertainty of the model. However, uncertainty of wheel odometry models is described in some detail by \cite{ari2017}, and is applicable here.

%E.g. $\mathbf{u}^v = [u^v_x, u^v_y, u^v_z]^\top$ indicates a vector in the vehicle coordinate system, and similarly $\mathbf{v}^w~=~[v^w_x, v^w_y, v^w_z]^\top$ indicates a vector in the world coordinate system. Scalars with ambiguity carry the superscript, but scalars that represent distance (e.g. $r$) are independent of coordinate system, so don't carry the subscript. Matrices use two superscripts to denote the coordinate systems in which it represents the transform e.g. $^v\mathbf{R}^w$ represents a rotation transform from vehicle to world coordinate system, and thus $^w\mathbf{R}^v = {^v}\mathbf{R}{^w}{^{\top}}$.

%\subsection{Planar odometry}

\subsection{Heading angle}
The heading angle $\theta(t)$ (with radians as a unit, for example) at any point in time $t$ (with seconds as a unit, for example) is given by integrating the continuous yaw rate function:
\begin{equation}
\theta(t_1) = \int_{0}^{t_1} \theta'(t) dt
\end{equation}
%Hence, the constant of integration is the absolute heading angle in some external world coordinate system (e.g. latitude and longitude). %This of course cannot be extracted without an external signal (e.g. GPS). 
%If we ignore 
Ignoring the constant of integration, we get absolute heading in the coordinate system of the position of the vehicle at time zero (i.e. the power on of the vehicle, or the start of running of the piece of implemented software). In the general case, we do not have the underlying yaw rate function, but rather we only have samples from the yaw rate sensor between the two times, $t_1$ and $t_2$. Thus, we can accumulate iteratively:
\begin{equation}
\theta(t_{2}) = \left(\int_{t_1}^{t_2} \theta'(t) dt\right) + \theta(t_{1})
\end{equation}
As the sensors are sampled (i.e. the continuous function is not available in reality), this is approximated as

\begin{equation}
\Delta \theta = \frac{\theta'_1 + \theta'_2}{2}\left(t_2 - t_1\right), \hspace{1cm}
\theta(t_2) \approx \theta_{2} = \Delta \theta + \theta_{1}
\label{eqn:deltTheta}
\end{equation}
where $\theta'_1$ and $\theta'_2$ are the yaw rate samples (for example, with radians per second as a unit) at the times $t_1$ and $t_2$ respectively, and are sampled approximations of the continuous function $\theta'(t)$. Estimating $\Delta \theta$ is done by taking the average of the two yaw rate samples (in rad/s) and multiplying by the equivalent time difference to get the heading angle (in rad). This is iterative, as a new sample arrives $\theta_{2}$ is assigned to the previous sample, $\theta_{1}$, and $\theta_{2}$ takes the value of the new sample. Thus from the yaw rate sensor, we can extract the absolute heading angle $\theta_t$ at any sample time $t$, and the delta heading angle from the previous sample $\Delta \theta$.% $\theta(t) = \int \theta'(t)$ is the continuous heading angle function, and $\Delta \theta = \theta_1 - \theta_2$ is an approximation of the difference at two points in time, $t_1$ and $t_2$ (with associated samples $\theta'_1$ and $\theta'_2$), obtained from the samples of the yaw rate sensor.

\subsection{Planar Displacement}

\begin{figure}
    \centering
    \begin{subfigure}[b]{0.2\textwidth}
        \centering
        \includegraphics[height=3cm]{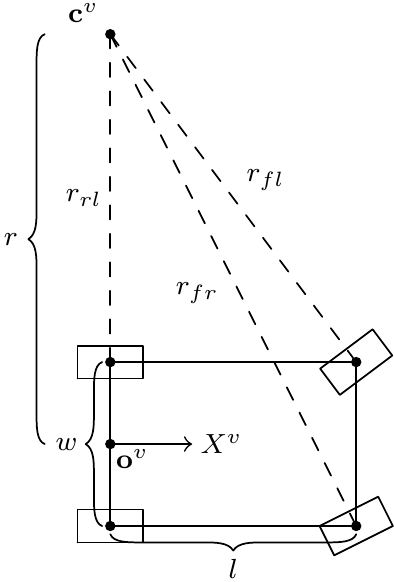}
        \caption{}
    \end{subfigure}
    \begin{subfigure}[b]{0.2\textwidth}
        \centering
        \includegraphics[height=3cm]{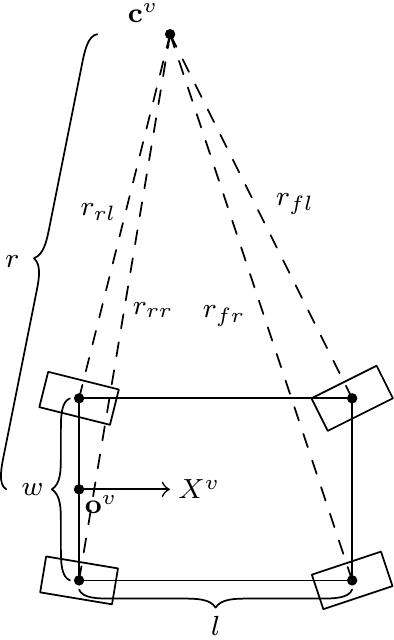}
        \caption{}
    \end{subfigure}
    \caption{Odometry models: a) Fixed rear steering, and b) adaptive rear steering. All points on the vehicle turn through a fixed instantaneous turning centre ($\mathbf{c}^v$). $w$ is the distance between wheel pairs, and $l$ is the wheelbase.}
    \label{fig:ackerman1}
    \vspace{-0.7cm}
\end{figure}

Planar odometry has an instantaneous centre of rotation \citep{jazar2008} as shown in Figure \ref{fig:ackerman1}. The integration time is short enough (\textasciitilde10-20 ms on a system bus) to consider the curvature to be constant between two samples. The vehicle, in the two dimensions of the plane, can be considered to be rigidly moving. % That is,
The relative positions of the points of contact of the tyres with the surface of the road remain constant. Hence, if the vehicle moves between two points in time, $t_1$ and $t_2$, and the angle, $\Delta \theta$, then the distance moved of any point is

\begin{equation}
d = r \Delta \theta
\label{eqn:dist}
\end{equation}
where $r$ is the distance of the point on the vehicle to the instantaneous centre of rotation. $\Delta \theta$ is given by (\ref{eqn:deltTheta}). Given a set of four samples of $d$ for the four wheels of the vehicle $\left\{d_{rl},d_{rr},d_{fl},d_{fr}\right\}$, the estimate of the distance from the wheel position to the turning centre (Figure \ref{fig:ackerman1}) is given by 
\begin{align}
r_{i} & = \frac{d_{i}}{\Delta \theta}, \hspace{0.5cm} i \in \left\{  rl, rr, fl, fr \right\} \label{equ:ri}
\end{align}
For the case with fixed rear steering (Figure \ref{fig:ackerman1}(a)), we can then get four estimates of the distance of the vehicle datum to the turning centre, with the average being our final estimate. %The averaging in works, as we are just finding the distance of the instantaneous rotation centre along the $X^v$-axis of the vehicle, assuming no rear wheel steering. That is, we are only solving for a single variable.
\begin{equation}
    r_{1} = r_{rl} - w/2, \hspace{0.5cm}
    r_{2} = r_{rr} + w/2, \hspace{0.5cm}
    r_{3} = \sqrt[+]{r_{fl}^2 - l^2} - w/2, \hspace{0.5cm}
    r_{4} = \sqrt[+]{r_{fr}^2 - l^2} + w/2
\end{equation}
\begin{equation}
    r = \frac{r_{1} + r_{2} + r_{3} + r_{4}}{4} \label{eqn:avgRadii}
\end{equation}
The distance from the centre of motion to the datum $r$ is estimated using the average of the four extracted radii. $w$ is the distance between the wheels, and $l$ is the length from the front wheel pair to the rear wheel pair (i.e. wheelbase). The yaw rate is signed to give the ``left'' or ``right'' motion of the vehicle, and the wheel distances $\left\{d_{rl},d_{rr},d_{fl},d_{fr}\right\}$ are signed to give the ``forward'' or ``backward'' motion of the vehicle. The instantaneous centre of rotation is therefore $\mathbf{c}^v=[0, r, 0]^\top$. In the case of adaptive rear steering (Figure \ref{fig:ackerman1}(b)), we have two free parameters for $\mathbf{c}^v = [c^v_x, c^v_y, 0]^\top$. We solve this using least squares. The error function is given by
\begin{equation}
    E (\mathbf{c}^v) = \sum_i\left|\mathbf{w}^v_i - \mathbf{c}^v\right|^2_2 - r_i^2, \hspace{0.5cm} i \in \left\{rl, rr, fl, fr\right\}
\end{equation}
and solving the partial differential equations $\delta E (\mathbf{c}^v) / \delta c^v_x = 0$ and $\delta E (\mathbf{c}^v) / \delta c^v_y = 0$ to obtain the estimate for $\mathbf{c}^v$. %In this case, 
$\mathbf{w}^v_i$ indicates the position, in the vehicle coordinate system, of each of the wheels of the vehicle, given by appropriate combinations of $w$ and $l$. $r_i$ is from (\ref{equ:ri}). Given the estimate of $\mathbf{c}^v$, the datum distance is simply 
\begin{equation}
    r = |\mathbf{c}^v|_2            \label{eqn:rearSteerR}
\end{equation}
The motion vector, in vehicle coordinates, is given by 
\begin{equation}
    \Delta \mathbf{p}^v = \left[r\sin{\Delta \theta}, r\cos{\Delta \theta}, 0\right]^\top
\end{equation}
where $r$ is estimated from (\ref{eqn:avgRadii}) or (\ref{eqn:rearSteerR}) as appropriate, and $\Delta \theta$ is estimated from (\ref{eqn:deltTheta}). Given the heading angle, $\theta_{1}$, at time $t_1$, the overall position of the vehicle at a given time $t_2$ is given by the accumulation
\begin{equation}
    \mathbf{p}_{2}^w = {^v}\mathbf{R}^{w}_{1}\Delta\mathbf{p}^v + \mathbf{p}_{1}^w
    \label{eqn:vehPos}
\end{equation}
where ${^v}\mathbf{R}^w_{1}$ is the $3 \times 3$ rotation matrix equivalent of the heading angle, $\theta_{1}$, the rotation about the $Z^w$-axis. This is accumulative, so in the next iteration of the odometry calculation, $\mathbf{p}_{2}^w$ is assigned to $\mathbf{p}_{1}^w$.

% This is demonstrated in Figure \ref{fig:ackerman2}.

%\begin{figure}[!hbp]
%\vspace{-0.4cm}
%    \centering
%    \includegraphics[width=0.3\textwidth]{images/ackerman2.pdf}
%    \caption{Position and orientation of the vehicle in the world coordinate system. As it moves, all points on the vehicle move through a common angle ($\Delta \theta$) about a common instantaneous centre of rotation ($\mathbf{c}^w$).}
%    \label{fig:ackerman2}
%\end{figure}

\subsection{Suspension model}

\begin{figure}
    \centering
    \begin{subfigure}[b]{0.25\textwidth}
        \centering
        \includegraphics[height=2.5cm]{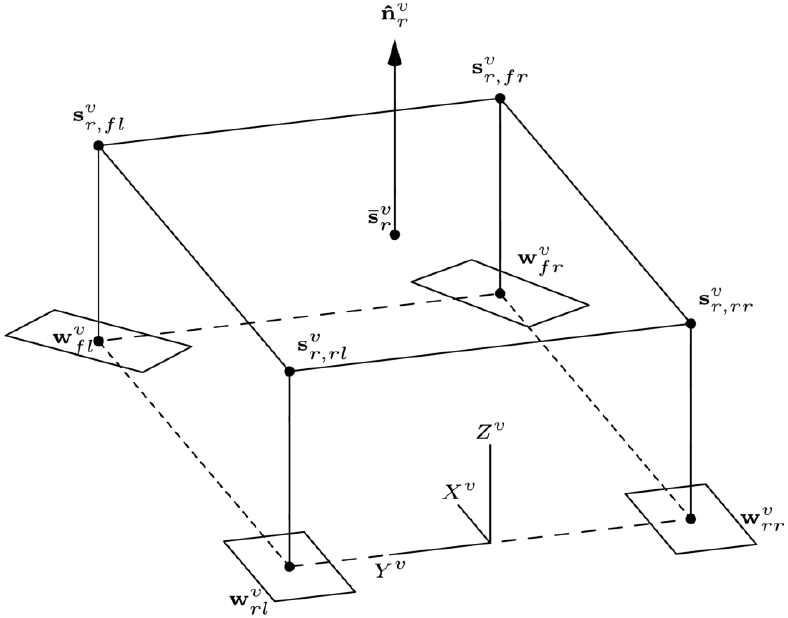}
        \caption{}
    \end{subfigure}
    \begin{subfigure}[b]{0.25\textwidth}
        \centering
        \includegraphics[height=2.5cm]{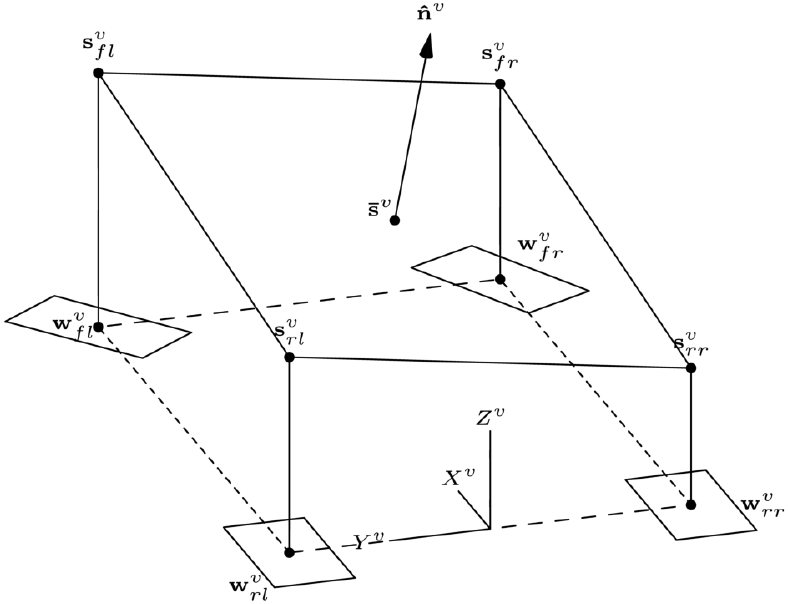}
        \caption{}
    \end{subfigure}
    \caption{The suspension plane models. a) settled state, and b) with loading}
    \label{fig:suspension}
    \vspace{-0.5cm}
\end{figure}

A sensor (e.g. a camera) located on a vehicle has a particular set of extrinsic calibration parameters (rotation and translation) in the vehicle coordinate system. %It is very common to assume that we can directly derive the orientation and height from the ground from the calibration parameters. 
Note, calibration is usually done against the rigid coordinate system of the vehicle body, which doesn't take into account the pitching, rolling and settling of the vehicle suspension. $\mathbf{s}_i^v$ is the suspension point in the settled state. $\mathbf{s}_i^v$ is obtained by taking the wheel positions $\mathbf{w}_i^v$, and setting the $z$ component to the height obtained from the calibrated linear potentiometers (Figure \ref{fig:suspension}(a)). That is, if we set $h_i$ as the set of heights from the sensors, then $\mathbf{s}_i^v = [w_{i,x}^v, w_{i,y}^v, h_i]^\top$. With no load on the vehicle, or no acceleration, the suspension will be in a settled state. The points form a plane in the vehicle coordinate system, defined by a normal vector $\mathbf{\hat{n}}_{r}^v$ and a reference point $\mathbf{\bar{s}}_r^v = \frac{1}{4}\sum \mathbf{s}_i^v$, which can be obtained using ordinary least squares. In all cases above, $i \in \{rl, rr, fl, fr\}$. In live operation, the suspension will change (Figure \ref{fig:suspension}(b)). We can use the exact same procedure to extract a live description of the suspension plane model with the normal vector $\mathbf{\hat{n}}^v$ and reference point $\mathbf{\bar{s}}^v$. Only the $z$ component of $\mathbf{\bar{s}}^v$ will be different compared to  $\mathbf{\bar{s}}_r^v$, as the positions of the wheels $\mathbf{{w}}_i^v$ do not change with suspension changes. In order to combine the suspension changes with the odometry, we wish to represent it as a rotation matrix $^v\mathbf{R}^v_s$ and a translation vector $\mathbf{t}^v_s$. The translation is straightforwardly
~
~
%The previous paragraph describes a calibration procedure in which the reference data ($\mathbf{\hat{n}}_{r}^v$ and $\mathbf{\bar{s}}_{r}^v$) is extracted. 
~
\begin{equation}
    \mathbf{t}^v_s = \mathbf{\bar{s}}^v_r - \mathbf{\bar{s}}^v
\end{equation}
%Obtaining 
The rotation matrix %from the two normal vectors is well known. It 
is given by the axis-angle formula (recalling that $\mathbf{\hat{n}}^v$ and $\mathbf{\hat{n}}_r^v$ are both unit vectors):
\begin{equation}
    \mathbf{a} = [a_x, a_y, a_z]^\top = \mathbf{\hat{n}}^v \times \mathbf{\hat{n}}^v_r \nonumber, \hspace{0.5cm}
    s = |\mathbf{a}|, \hspace{0.5cm}
    c = \mathbf{\hat{n}}^{v} \cdot \mathbf{\hat{n}}^{v}_{r}
\end{equation}
where $s$ and $c$ are the sine and cosine of the angle between $\mathbf{\hat{n}}^v$ and $\mathbf{\hat{n}}_r^v$. Then
\scriptsize
\begin{equation}
    ^v\mathbf{R}^v_s \! = \! \!
    \left[ \! \!
        \begin{array}{ccc}
            c + a_x^2(1-c) & a_x a_y (1 - c) - a_z s & a_x a_z (1 - c) + a_y s \\
            a_x a_y(1 - c) + a_z s & c + a_y^2(1 - c) & a_y a_z (1 - c) - a_x s\\
            a_x a_z(1 - c) + a_y s & a_y a_z (1 - c) + a_x s & c + a_z^2 ( 1 - c) 
        \end{array}
    \! \! \right] \nonumber
\end{equation}
\normalsize
Some notes on the assumptions of this model. Firstly, points on the vehicle body that are planar will remain planar under different suspension configurations. While there can be some flex in vehicle body, for the most part it can be considered rigid, and thus this assumption is valid. Secondly, the different suspension configurations cause our reference points to move vertically. Actually, this is not the case, as a changing suspension will cause a rotation of the vehicle body. However, vertical motion will dominate over lateral motion, and thus we can ignore the lateral motion of the reference points. %A related assumption is that
%Lastly, suspension will cause a rotation only in roll and pitch, and that yaw rotation is negligible. Again, this is a valid assumption, as a yaw would be equivalent to a ``twist'' of the vehicle body, which does not happen with a real vehicle, or at least only happens to a negligible degree.

\subsection{Sensor calibration}

%In the following section, we describe now how to obtain the sensor position in the world coordinate system, given the previously described odometry and suspension motion estimations. It is pertinent to discuss the calibration of such sensors. 
Sensors on the vehicle have an extrinsic calibration represented as a rotation matrix $^v\mathbf{R}^c_e$ and a position vector $\mathbf{c}^v_e$, in vehicle coordinates %.We do not discuss how the calibration is achieved, as there is already a wealth of work in the area 
\citep{, choi2018}.
%rokita2012, hold2009}. 
Typically, the calibration procedure extracts the $^v\mathbf{R}^c_e$ and $\mathbf{c}^v_e$ of the cameras against an external reference, such as a local road plane defined approximately by the points of contact of the 4 wheels with the ground, or ground markings on such a surface, taking into account only a nominal reference suspension. This leads to a definition of the sensor calibration against the road plane rather than against the vehicle body. Thus, if the calibration runs when there is a heavy loading in the vehicle, causing a suspension change from the nominal, then the system will not calibrate for a ``true'' extrinsic position. To solve this, one must account for the suspension during the calibration procedure, and this gives us the extrinsic camera parameters considering the “nominal” or reference suspension. %A means to do this is as follows. 
As described previously, we can get the rotation and translation due to suspension changes from the nominal ($^v\mathbf{R}^v_{s}$, $\mathbf{t}^v_{s}$). During the calibration procedure, we get a calibrated rotation and translation ($^v\mathbf{R}^c_{cal}$, t$_{cal}$). However, these include the offsets due to the suspension, as the algorithm runs when the suspension is different from nominal. To get the true extrinsic camera positions:%, we must take into account the suspension
\begin{equation}
    ^v\mathbf{R}^c_e = {^v}\mathbf{R}^{v\top}_s \: {^v}\mathbf{R}^c_{cal}
\end{equation}
and similarly for the calibrated camera position $\mathbf{c}^v_e$. This is done for each camera, and then gives the calibration against the nominal or reference suspension setting for each camera.

\subsection{Combining motions}

The overall pose of the camera in the vehicle coordinate system is therefore given by the composition of the suspension model and calibration rotations
\begin{equation}
    ^v\mathbf{R}^c_{p} = {^v}\mathbf{R}^c_e \: {^v}\mathbf{R}^v_s
\end{equation}
The position of the camera in vehicle coordinates can be given by (note $\mathbf{t}^v_s = [0, 0, h]^\top$):
\begin{equation}
    \mathbf{c}^v_{p} = {^v}\mathbf{R}^v_s ( \mathbf{c}^v_e + \mathbf{t}^v_s)
\end{equation}
The position of the sensor in the world coordinate system can then be given by
\begin{equation}
    {^w}\mathbf{R}^c_{p} = {^v}\mathbf{R}^c_{p} \: ^w\mathbf{R}^v, \hspace{0.5cm}
    \mathbf{c}_p^w = {^v}\mathbf{R}^w \mathbf{c}^v_{p} + \mathbf{p}^w
\end{equation}
with $^w\mathbf{R}^v$ obtained from the odometry heading angle (\ref{eqn:deltTheta}), and $\mathbf{p}^w$ is the vehicle position from equation (\ref{eqn:vehPos}). %And thus, we have a full description of the pose of the sensor in the world coordinate system.

\section{Results}

\label{sec:results}

Ground truth isn't available for 2.5D odometry, or for suspension in general, as DGPS is the only ground truth sensor available in our system. With that in mind, we compare the planar 2D odometry to the ground truth DGPS, and then subjectively examine the performance of 2.5D odometry in the context of two vision applications; visualisation (top view) and computer vision (motion segmentation).

% As there is no ground-truth available for the 2D odometry we therefore analyse and present our results in three separate sections, focusing on: Planar Odometry, Suspension Sensor Behaviour, Visualisation, and Computer Vision.

\subsection{Planar Odometry}

\begin{figure}
    \centering
    \begin{subfigure}[b]{0.4\textwidth}
        \centering
        \includegraphics[height=2cm]{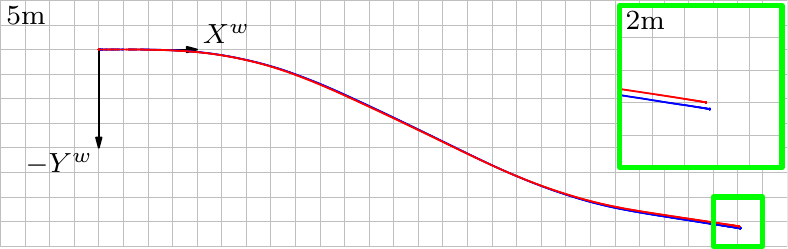}
        \caption{}
    \end{subfigure}
    \begin{subfigure}[b]{0.4\textwidth}
        \centering
        \includegraphics[height=2cm]{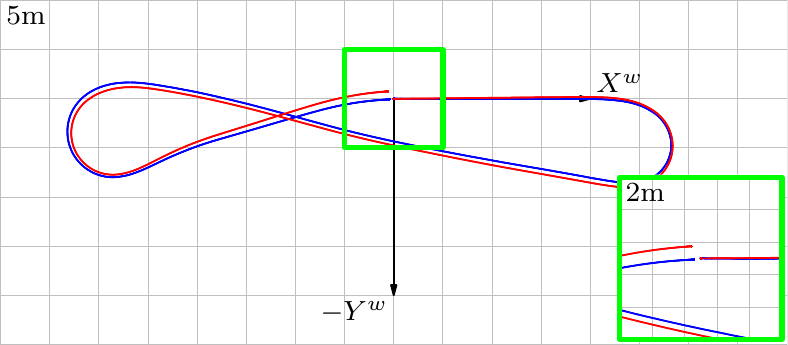}
        \caption{}
    \end{subfigure}
    \caption{Calculated vehicle position (\textcolor{red}{red}) versus ground truth (DGPS) (\textcolor{blue}{blue}): ``standard'' manoeuvre (a) and complex manoeuvre (b).}
    \label{fig:graph_odom1}
\end{figure}

Figure \ref{fig:graph_odom1} show the trajectory from two vehicle manoeuvres. The error for the simpler manoeuvre at the end of the trajectory is $0.46$ m (Figure \ref{fig:graph_odom1}(a)), whereas the more complex manoeuvre has an overall drift of only $0.82$ m (Figure \ref{fig:graph_odom1}(b)). Hence, this shows that 2D planar odometry is a sufficiently accurate input to our model.

\subsection{Suspension Sensor Behaviour}

\begin{figure}
    \centering
    \begin{subfigure}[b]{0.2\textwidth}
        \centering
        \includegraphics[height=2.3cm]{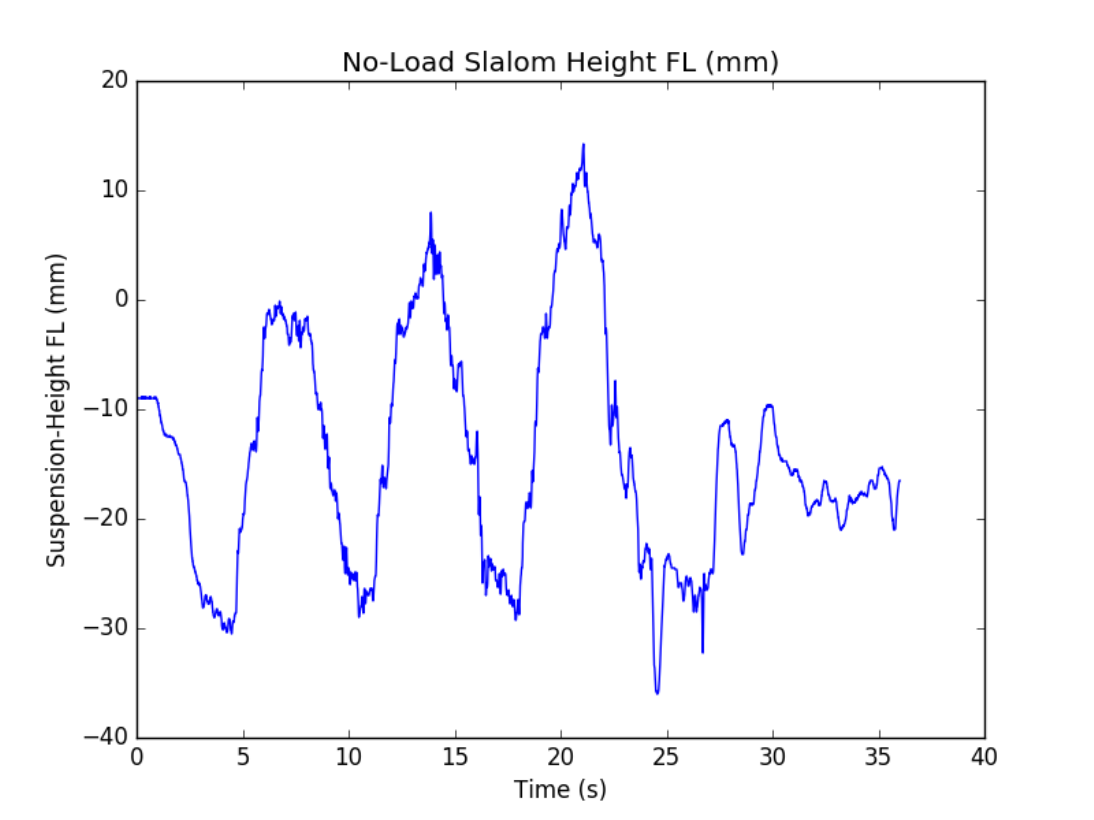}
        \caption{front left}
    \end{subfigure}
    \begin{subfigure}[b]{0.2\textwidth}
        \centering
        \includegraphics[height=2.3cm]{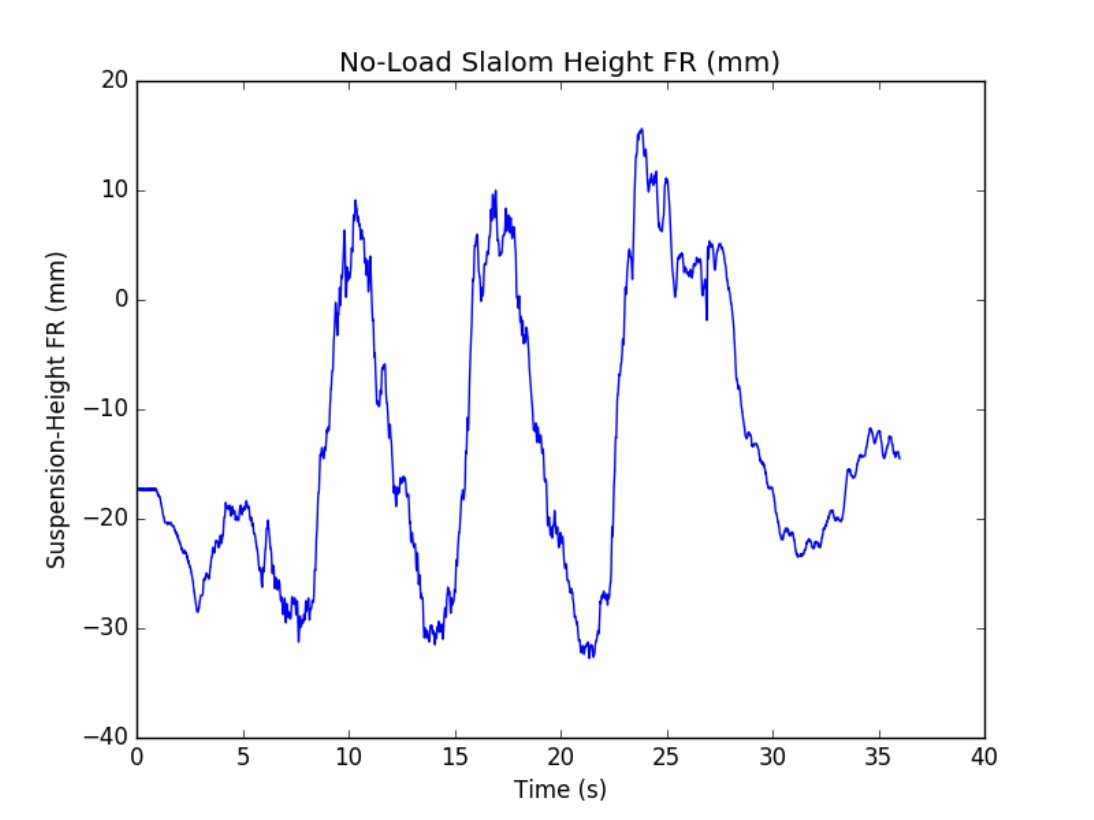}
        \caption{front right}
    \end{subfigure}
    \begin{subfigure}[b]{0.2\textwidth}
        \centering
        \includegraphics[height=2.3cm]{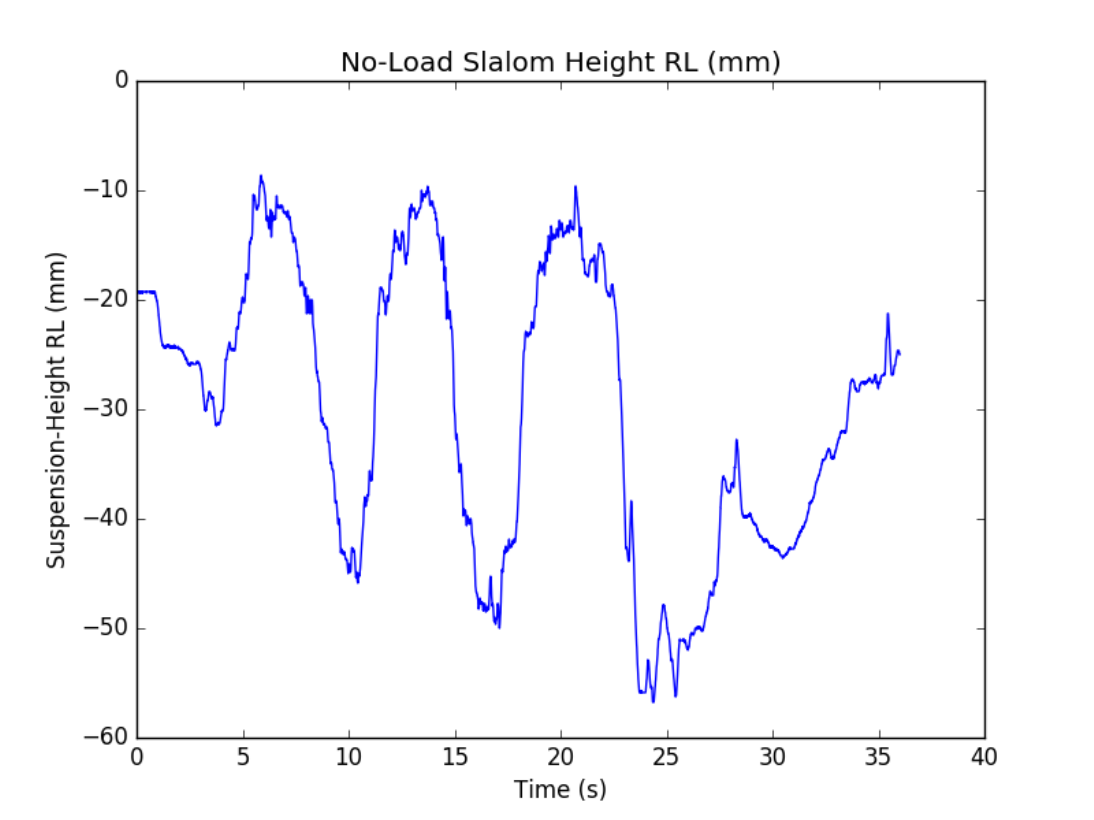}
        \caption{rear left}
    \end{subfigure}
    \begin{subfigure}[b]{0.2\textwidth}
        \centering
        \includegraphics[height=2.3cm]{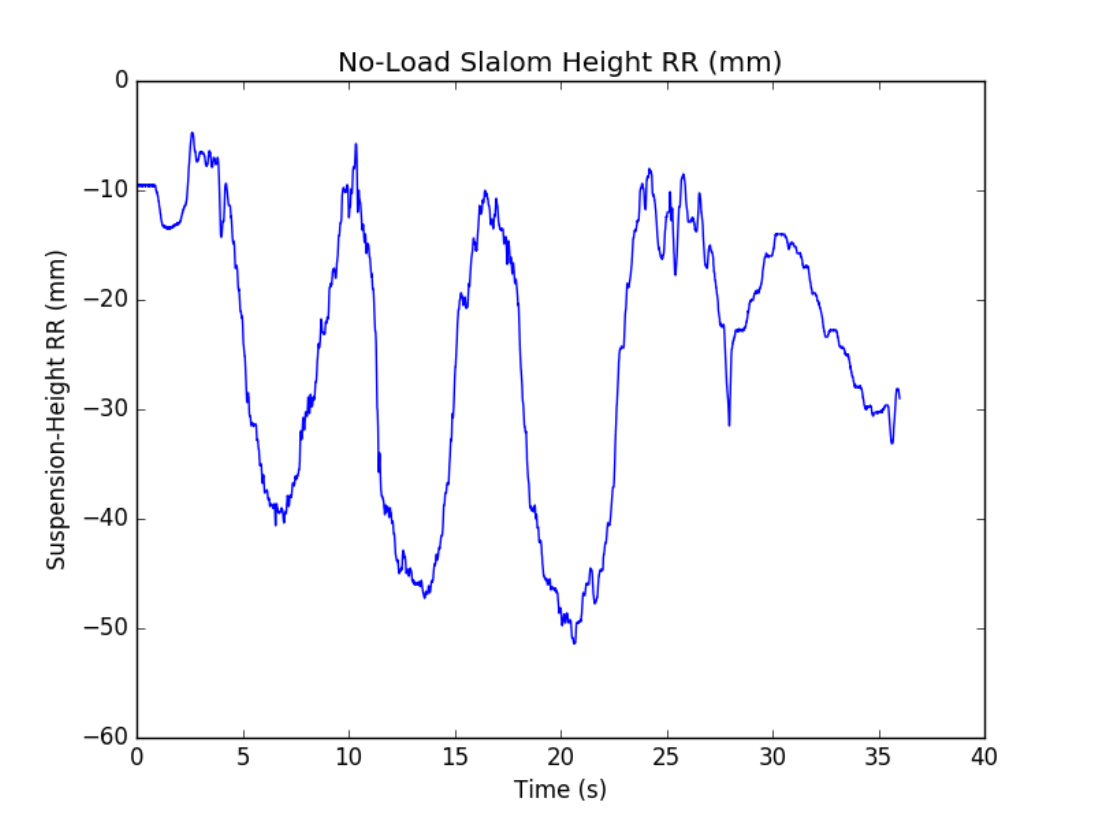}
        \caption{rear right}
    \end{subfigure}
    \caption{Plots of the suspension sensor height as a function of time for slalom motion.}
    \label{fig:slalom}
    \vspace{-0.2cm}
\end{figure}

\begin{figure}
    \centering
    \begin{subfigure}[b]{0.2\textwidth}
        \centering
        \includegraphics[height=2.3cm]{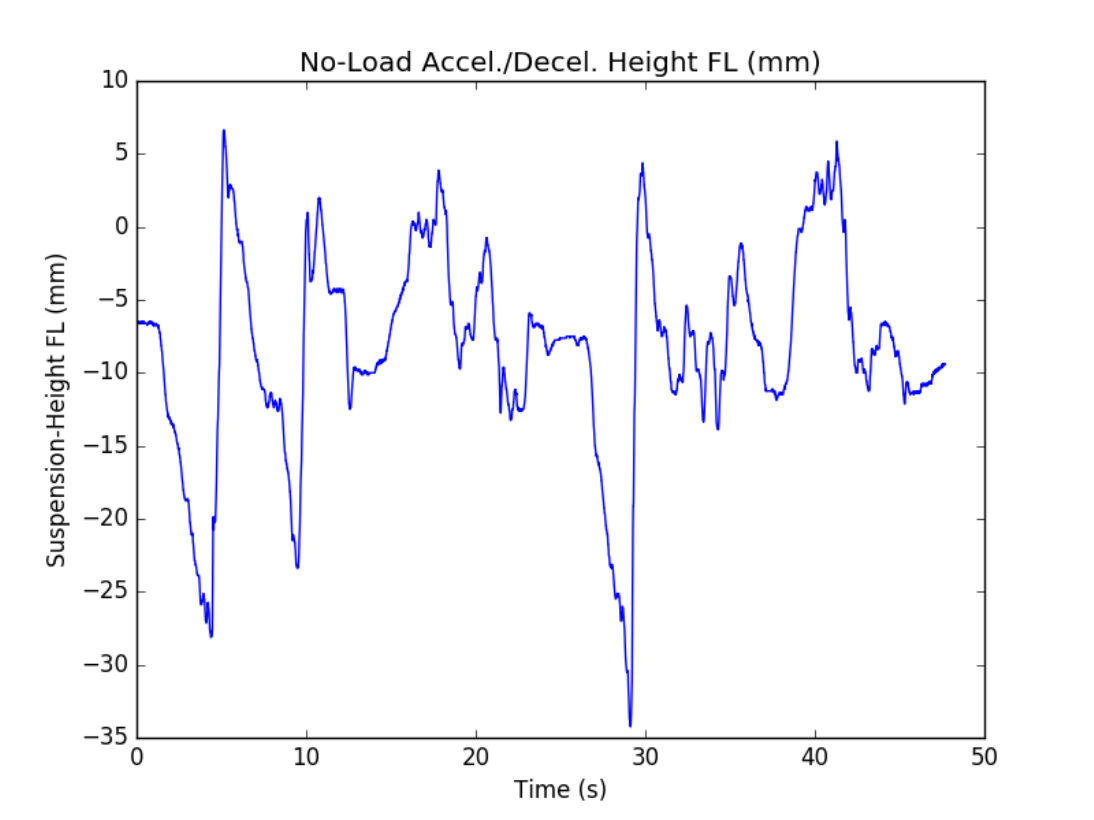}
        \caption{front left}
    \end{subfigure}
    \begin{subfigure}[b]{0.2\textwidth}
        \centering
        \includegraphics[height=2.3cm]{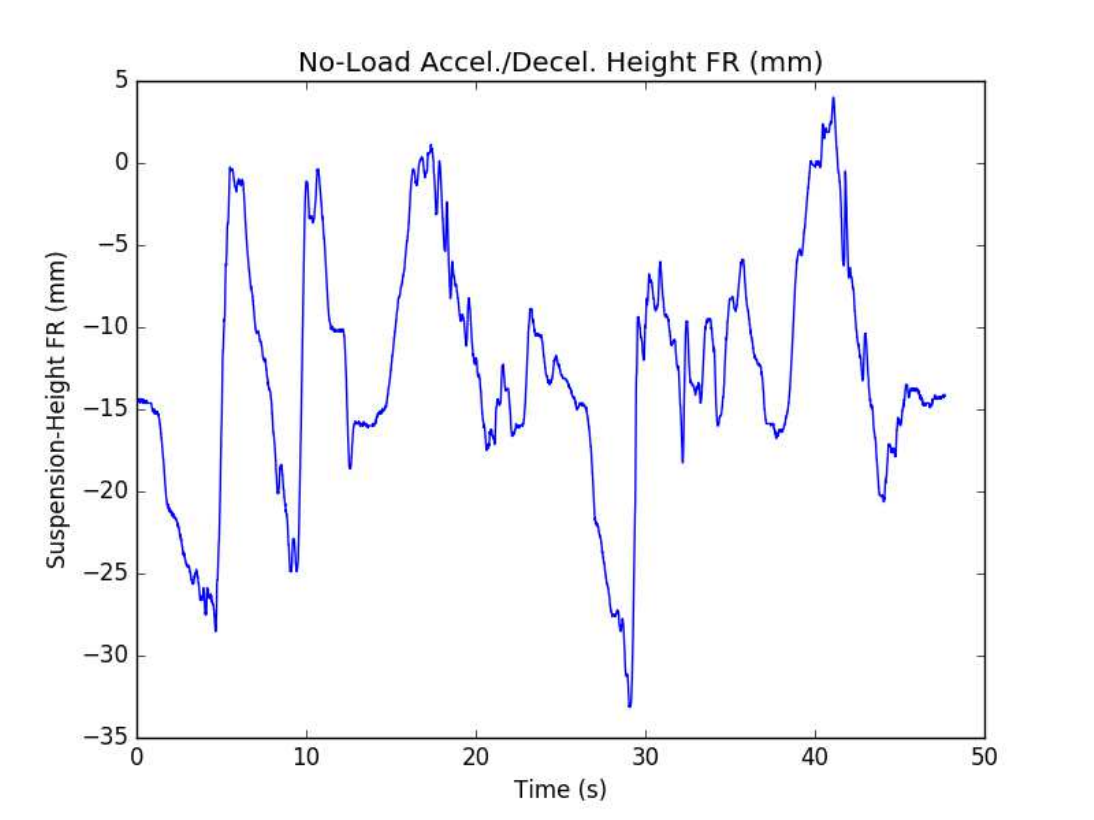}
        \caption{front right}
    \end{subfigure}
    \begin{subfigure}[b]{0.2\textwidth}
        \centering
        \includegraphics[height=2.3cm]{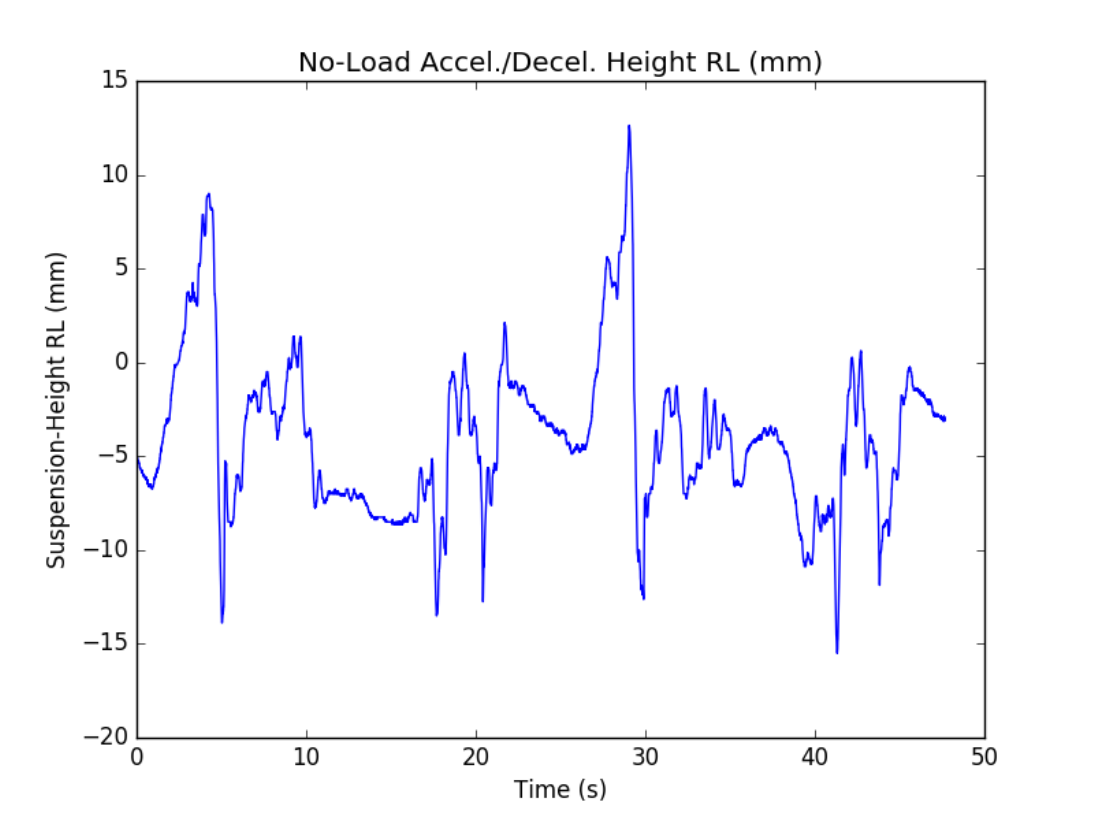}
        \caption{rear left}
    \end{subfigure}
    \begin{subfigure}[b]{0.2\textwidth}
        \centering
        \includegraphics[height=2.3cm]{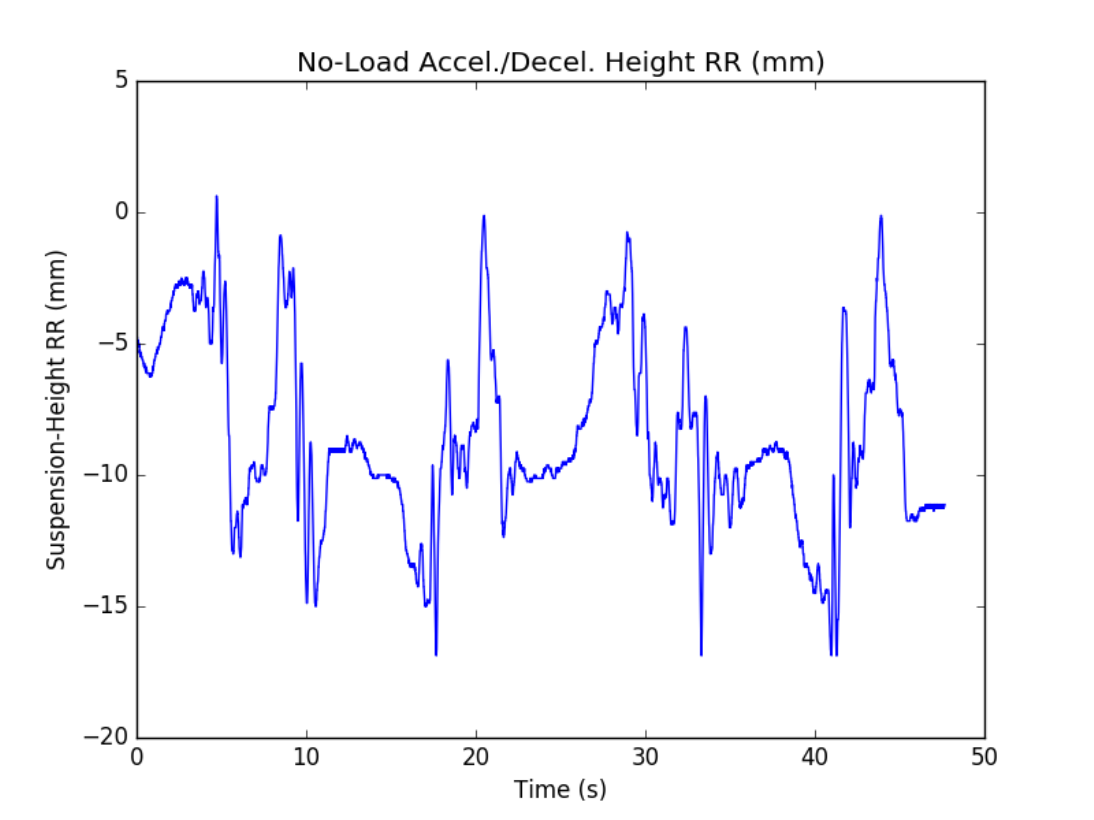}
        \caption{rear right}
    \end{subfigure}
    \caption{Plots of the suspension sensor height as a function of time for acceleration and deceleration.}
    \label{fig:accel_1}
\end{figure}

Experiments were performed to test the accuracy of the algorithm that compensates calibration based on the suspension. The first experiment involved checking the stability and accuracy of the input ego-vehicle suspension data. The input data comes from sensors mounted at the arches of the four wheels: Front-left ($fl$), Front-Right ($fr$), Rear-Left ($rl$), and Rear-Right ($rr$). These sensors measure changes in their vertical height from the ground plane. Two particular cases were studied: slalom motion (driving in arcs or zig-zags), and acceleration and deceleration. %(starting and stopping) of the ego-vehicle
The data was recorded from the CAN bus of a test vehicle. Figures \ref{fig:slalom} \& \ref{fig:accel_1} show the heights (mm) of each of the wheel arch sensors plotted as a function of time. For the slalom motion it can be seen that the peaks and troughs of the plots of the suspension sensors mounted on the left and right side of the ego-vehicle were out of phase (Figure \ref{fig:slalom}) i.e. the peaks in the left wheel pair occur at the same moment in time as the troughs in the right wheel pair. This agrees with the physics of the use case, namely centripetal force. During the slalom the weight of the ego-vehicle is transferred directly to one side. Hence, the ego-vehicle becomes unbalanced with one side raised and the other side lowered. Similarly, for acceleration and deceleration (Figure \ref{fig:accel_1}) it was seen that the plots of the front and rear side are out of phase. Again, in agreement with the physics of the transfer of loading of the vehicle. During acceleration from rest the weight pushes down at the rear of the ego-vehicle and the front pitches upwards. Whereas during braking the opposite effect occurs.

\subsection{Visualisation}

\begin{figure}
    \centering
    \begin{subfigure}[b]{0.2\textwidth}
        \centering
        \includegraphics[height=2.5cm]{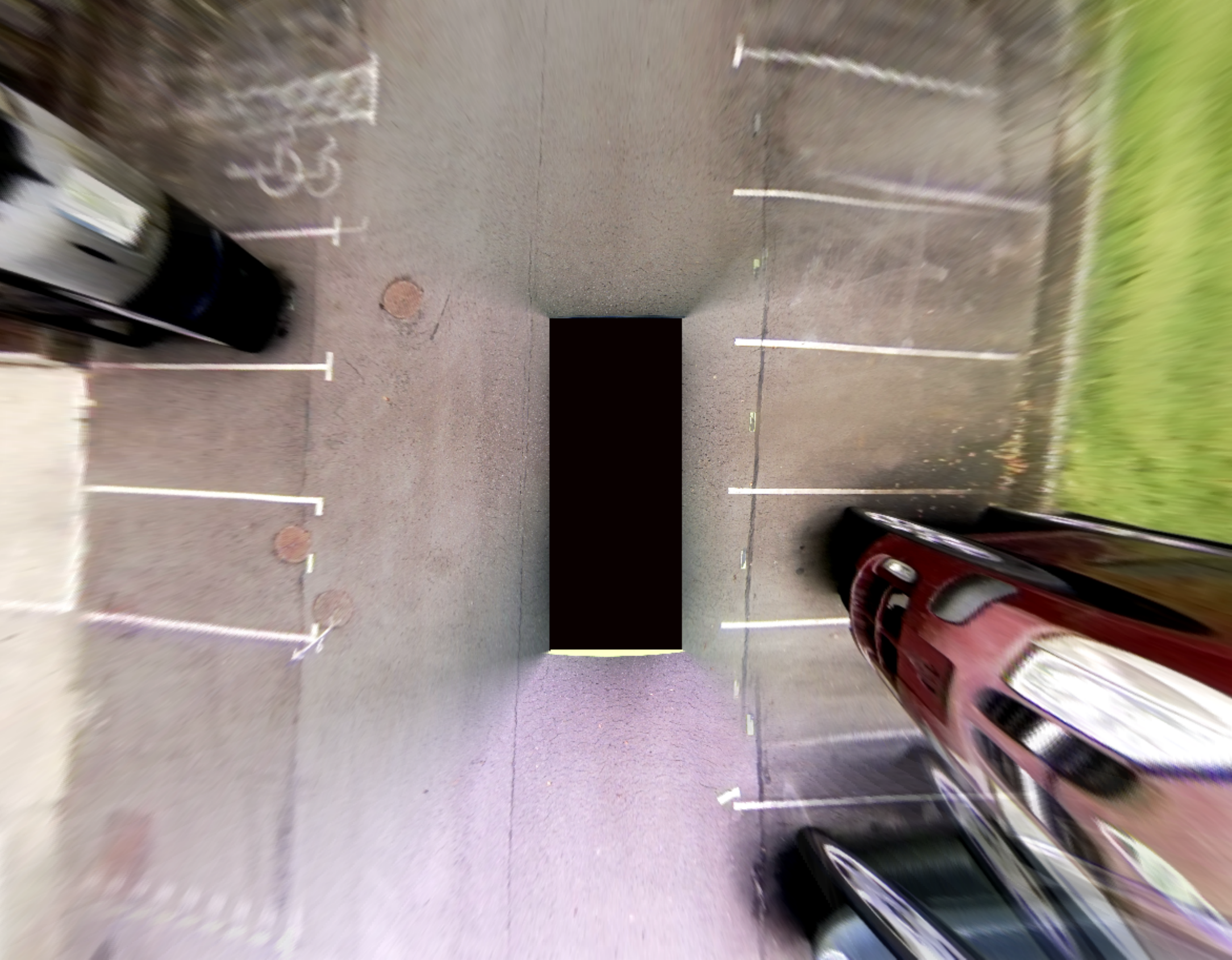}
        \caption{}
    \end{subfigure}
    \begin{subfigure}[b]{0.2\textwidth}
        \centering
        \includegraphics[height=2.5cm]{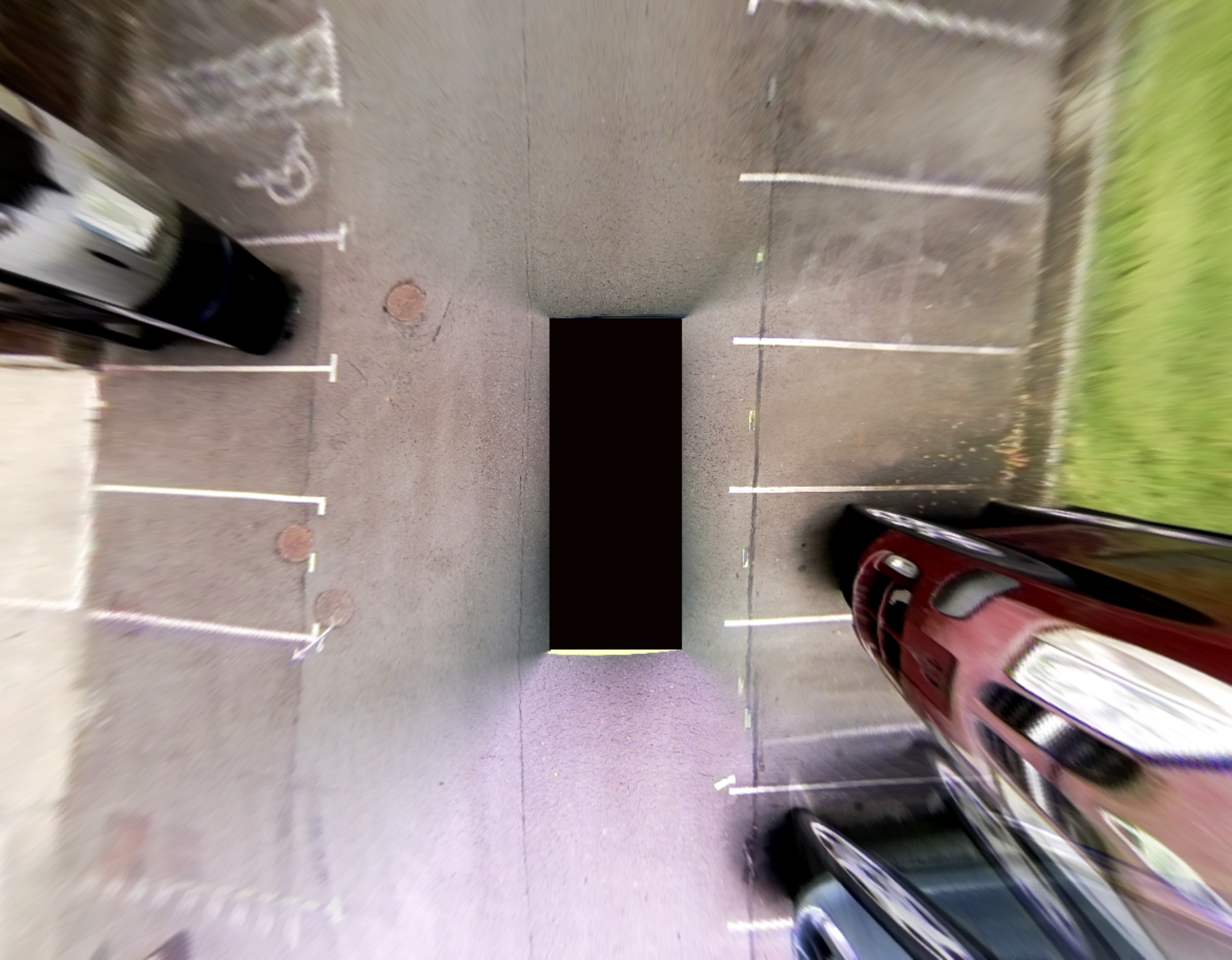}
        \caption{}
    \end{subfigure}
    \vspace{-0.5cm}
    \caption{Top-view images generated using calibrated extrinsics (a), and suspension-corrected extrinsics (b).}
    \label{fig:top_views}
    \vspace{-0.3cm}
\end{figure}

We generate a top-view of the vehicle's surroundings using four fisheye cameras on the vehicle: front, rear, and the two wing-mirrors. %Such a top-view is a typical product. 
The aim is to analyse the visual impact of utilising the suspension-corrected extrinsic parameters compared to the nominal extrinsic parameters. % during the types of motion described in the previous section. 
Table \ref{table:1} shows the nominal and suspension-compensated extrinsic calibration values for the scene shown in Figure \ref{fig:top_views}. The discontinuities between the parts of the images created by different cameras is quite evident in the top-view without suspension corrected extrinsic parameters. %Particular attention should be paid to the overlap regions of the cameras projection, the diagonal from the black square (ego vehicle) to the corner of the image, where the impact of extrinsic error can be observed best.  

\begin{table}
%\small
\fontsize{9}{8}\selectfont
\centering
  \begin{tabular}{|c|c|c|c|c||c|c|c|c|}
    \hline
    %\multirow{}{}{}
    &
      \multicolumn{4}{c||}{Nominal} &
      \multicolumn{4}{c|}{Suspension-Compensated}\\
      \hline
    & FV & RV & MVL & MVR & FV & RV & MVL & MVR \\
    \hline
     Height (mm) & 603.23    & 880.29    & 950.97  & 966.74 & 592.12 & 887.23    & 944.20 & 964.89\\
     \hline
     Rot. X ($^\circ$)   & 91.08 & 64.24 & 61.67  & 62.11 & 92.48 & 62.66     & 61.50   & 62.16\\
     \hline
     Rot. Z1 ($^\circ$)  & 89.96 & -90.71 & 167.94  & 2.87 & 91.23 & -89.58    & 168.03  & 3.96\\
     \hline
     Rot. Z2 ($^\circ$)  & -0.53 & 0.34  & 3.51   & -6.42 & -0.40 & 0.35      & 3.32    & -6.53\\ 
    \hline
  \end{tabular}
  \caption{Nominal and suspension-compensated extrinsic camera calibration values.}
\label{table:1}
\end{table}

\subsection{Computer Vision}

\begin{figure}[!h]
    \centering
    \includegraphics[width=.4\textwidth]{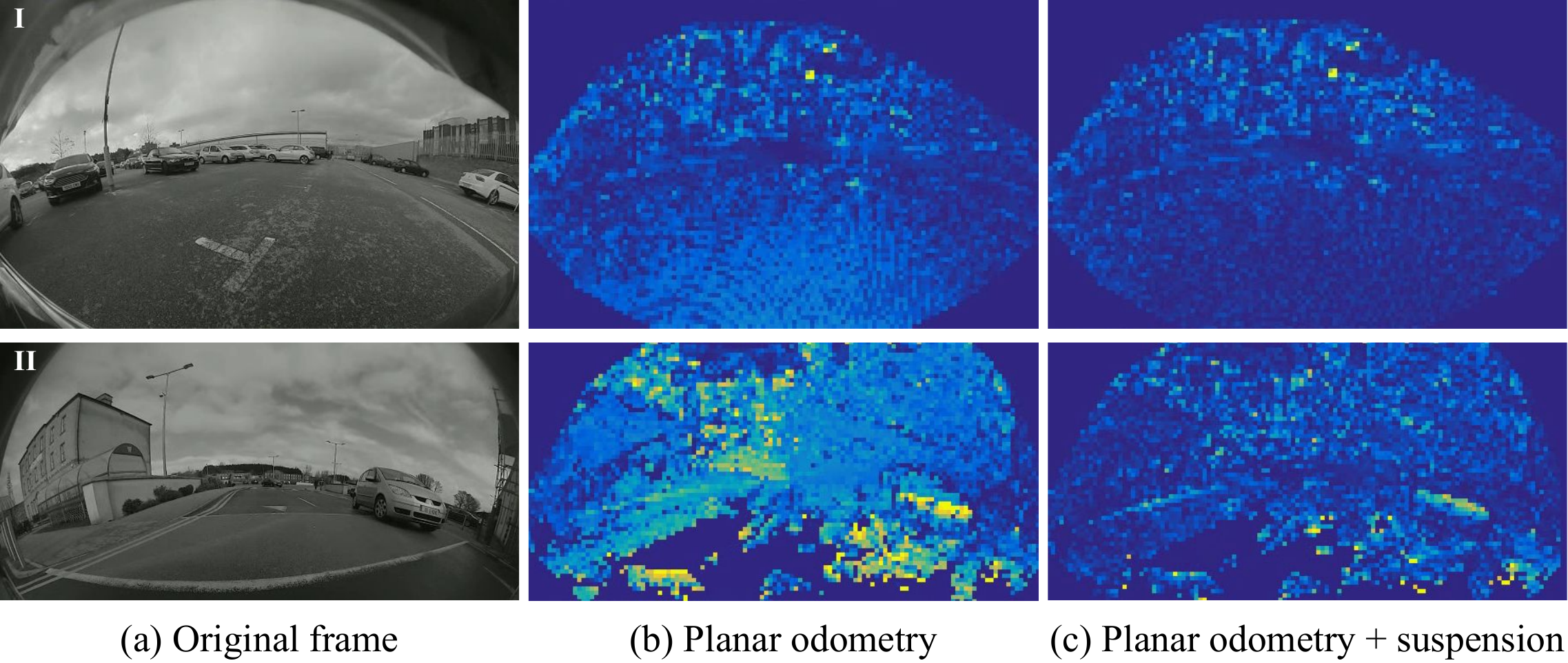}
    \caption{Motion segmentation maps: planar odometry (b), and planar odometry incorporating suspension (c). Motion likelihood increases from dark blue to red. The planar odometry results show that static objects are incorrectly predicting with much larger motion likelihoods since motion due to suspension is not compensated.}
    \label{fig:mod_results}
    \vspace{-0.5cm}
\end{figure}

\cite{mariotti2019} describe a geometric means of motion segmentation, and mention explicitly that the results in that paper are generated from a three degrees of freedom odometry, giving the position of the sensor in a world coordinate system. Here we briefly show some results of just using the planar odometry (Figure \ref{fig:mod_results}(b)) versus the planar odometry incorporating suspension sensors (Figure \ref{fig:mod_results}(c)). Figure \ref{fig:mod_results}(a) shows the original frame. In Row I, the vehicle is turning with rolling of the vehicle on the suspension. In Row II, the vehicle is accelerating heavily, showing significant pitching. In both cases, it can be seen that the error in the motion segmentation map is significantly lower when suspension is taken into account.

\section{Conclusion}

We have presented an odometry estimation algorithm using a set of sensors (yaw rate, wheel speed and suspension) commonly available on some modern, commercially available vehicles. It is computationally inexpensive, as the amount of data to process is minimal, but still provides significant improvement compared to just considering a planar odometry. This could be considered a 2.5D odometry, as it does not give a full 3D odometry (like from visual odometry) but it offers more than just the case of planar (2D) odometry. The results presented demonstrate that the integration error of the planar odometry is low. For visualisation applications, such as top-view, the use of the suspension sensors reduces stitching artefacts in the overlap regions between multiple cameras. The improved sensor extrinsic measure, relevant to all on board sensors, is key for perception and thus building precise environmental maps for automated and autonomous driving systems. For computer vision, the 2.5D odometry offers an advantage in the suppression of false positives, in the case that the computer vision requires an odometry input. Future work will consist of more rigorous experiments to determine the accuracy of the algorithm, and to test its use as an input to other computer vision applications. Visual-Inertial Odometry is an interesting area of development in robotics, in particular. Some further future work may be in integrating the 2.5D odometry with visual odometry, in the same way that low cost inertial sensors are integrated with visual odometry in Visual-Inertial Odometry. This would integrate the work presented in this paper entirely into a Visual SLAM environment.

%\section*{Acknowledgments}

%The authors would like to thank their employer for giving them the opportunity to investigate original research. Thanks also to Letizia Mariotti (Valeo) for providing the motion segmentation results presented.

%%%%%%%%%%%%%%%%%%%%%%%%
\appendix

\bibliographystyle{apalike}

\bibliography{references}

\end{document}